\def\year{2021}\relax
\documentclass[letterpaper]{article} 
\usepackage{aaai21}  
\usepackage{times}  
\usepackage{helvet} 
\usepackage{courier}  
\usepackage[hyphens]{url}  
\usepackage{graphicx} 
\usepackage{natbib}  
\usepackage{caption} 
\title{Fake it Till You Make it: Self-Supervised Semantic Shifts for Monolingual Word Embedding Tasks }

\author {

        Maur\'{i}cio Gruppi,\textsuperscript{\rm 1}
        Sibel Adal{\i},\textsuperscript{\rm 1}
        Pin-Yu Chen\textsuperscript{\rm 2} \\
}
\affiliations {
    \textsuperscript{\rm 1} Rensselaer Polytechnic Institute, NY, USA \\
    \textsuperscript{\rm 2} IBM Research, NY, USA \\
    gouvem@rpi.edu, 
    adalis@rpi.edu,
    pin-yu.chen@ibm.com
}

\usepackage{amsmath}
\usepackage{amsfonts}
\usepackage{amssymb}
\usepackage{multirow}
\usepackage[ruled, linesnumbered]{algorithm2e}
\usepackage{graphicx}
\usepackage{caption}
\usepackage{subcaption}
\usepackage{booktabs}

\usepackage{color}

\newcommand{\ourmethodD}{S4-D}
\newcommand{\ourmethodA}{S4-A}

\begin{document}

\maketitle

\begin{abstract}
 The use of language is subject to variation over time as well as across social groups and knowledge domains, leading to differences even in the monolingual scenario. Such variation in word usage is often called lexical semantic change (LSC). The goal of LSC is to characterize and quantify language variations with respect to word meaning, to measure how distinct two language sources are (that is, people or language models). Because there is hardly any data available for such a task, most solutions involve unsupervised methods to align two embeddings and predict semantic change with respect to a distance measure. To that end, we propose a self-supervised approach to model lexical semantic change by generating training samples by introducing perturbations of word vectors in the input corpora. We show that our method can be used for the detection of semantic change with any alignment method. Furthermore, it can be used to choose the landmark words to use in alignment and can lead to substantial improvements over the existing techniques for alignment.
 We illustrate the utility of our techniques using experimental results on three different datasets, involving words with the same or different meanings. Our methods not only provide significant improvements but also can lead to novel findings for the LSC problem.
\end{abstract}

\section{Introduction}
Language use is deeply rooted in the
social, cultural and historical context that shapes it. It has been
shown that meaning of words change over time, being pushed by cultural
and societal transformations, a phenomenon named \textit{semantic change}
\cite{schmidt1963ullmann}.  For example, the
English word \emph{awful} was used in the sense of the words {\em
  impressive} or {\em majestic} before the year 1800, while, in modern
English, it describes {\em
  something objectionable}. Language is also subject to variation across different
communities for many different
reasons~\cite{schlechtweg-etal-2019-wind}. 
For example, the word \emph{model} can be used to refer to the design or version of a product (as in model of a car), 
or it could be used to refer to a mathematical model in a scientific paper.

In this paper, we develop a novel self-supervised semantic shift (S4)
method to detect words with different meaning in two corpora from the
same language (monolingual).

As methods based on the distributional property of words have been
shown to be very effective in encoding semantic relationship between
words~\cite{hamilton2016diachronic, sagi2009semantic,  bamler2017dynamic, kulkarni2015statistically} or even biases and stereotypes~\cite{bolukbasi2016man, caliskan2017semantics}, the task of identifying semantic change between words
using word embeddings, such as Word2Vec or
FastText~\cite{mikolov2013distributed, bojanowski2017enriching} has
gained a great deal of popularity. This task is often a difficult one as
it involves unsupervised methods (e.g. learning embeddings, alignment and/or mapping of words). For
example, in the recent SemEval-2020 competition \cite{schlechtweg2020semeval}, the highest scores were at about 70\% accuracy on a binary classification task to predict occurrence of semantic change across time periods in several languages.
The main challenge stems from the unsupervised nature of the problem, as training data is rare or non-existing, and is highly dependent on the input corpora. This impacts multiple aspects of the task.
In particular, to compare the embedding matrices $A, B \in \mathbb{R}^{n \times d}$ of two
separate corpora $\mathcal{A}$ and $\mathcal{B}$, where $n$ is the size of the common vocabulary and $d$ is the embedding dimension,
one must first align
them to make them comparable, usually via an Orthogonal Procrustes
(OP) method. The goal of OP is to learn an orthogonal transform matrix $Q$ (i.e., $Q^TQ = I_d$, where $I_d$ is the d-dimensional identity matrix),
that most closely maps $A$ to $B$, namely, 
$Q^* = \arg \min_{Q:Q^TQ=I_d} \left\lVert AQ - B \right\rVert$.
It has been shown that this problem accepts a closed-form solution via Singular Value Decomposition (SVD) \cite{schonemann1966generalized}.
The fact that $Q$ is an
orthogonal matrix makes it so that $AQ$ is only subject to unitary
transformations such as reflection and rotation, preserving the inner
product between its word vectors. 
Words whose vectors are used in
the OP are called \emph{landmarks} (or anchors), these are the words
over which we will enforce proximity in the alignment by minimizing the distance of its vectors.

Any
landmark choice incorporates an initial assumption to the solution. 
An ideal solution to this problem would involve a set of landmark words that are semantically stable (i.e. words that have the same sense) across the input corpora. In the context of diachronic embeddings, where the embedding is learned from adjacent time slices of text, the assumption is that words will change only slightly from time $t$ to $t+1$, therefore, without loss of generality, all words can be used as landmarks.
We refer to the use of all common words as landmarks as the global alignment.
Global alignment may introduce undesirable similarity between word vectors of a word $w$ that is used in different senses across corpora. 
As a consequence of the orthogonal transformation, words that are supposed to be closely aligned will be distant. Wang et al. \cite{wang2019cross} refer to this problem, in the context of word translation, as \emph{oversharing}, that is, aligning more words than one should. 

\begin{figure}[t]
    \centering
    \begin{subfigure}{0.4\textwidth}
        \centering
        \includegraphics[width=1\textwidth]{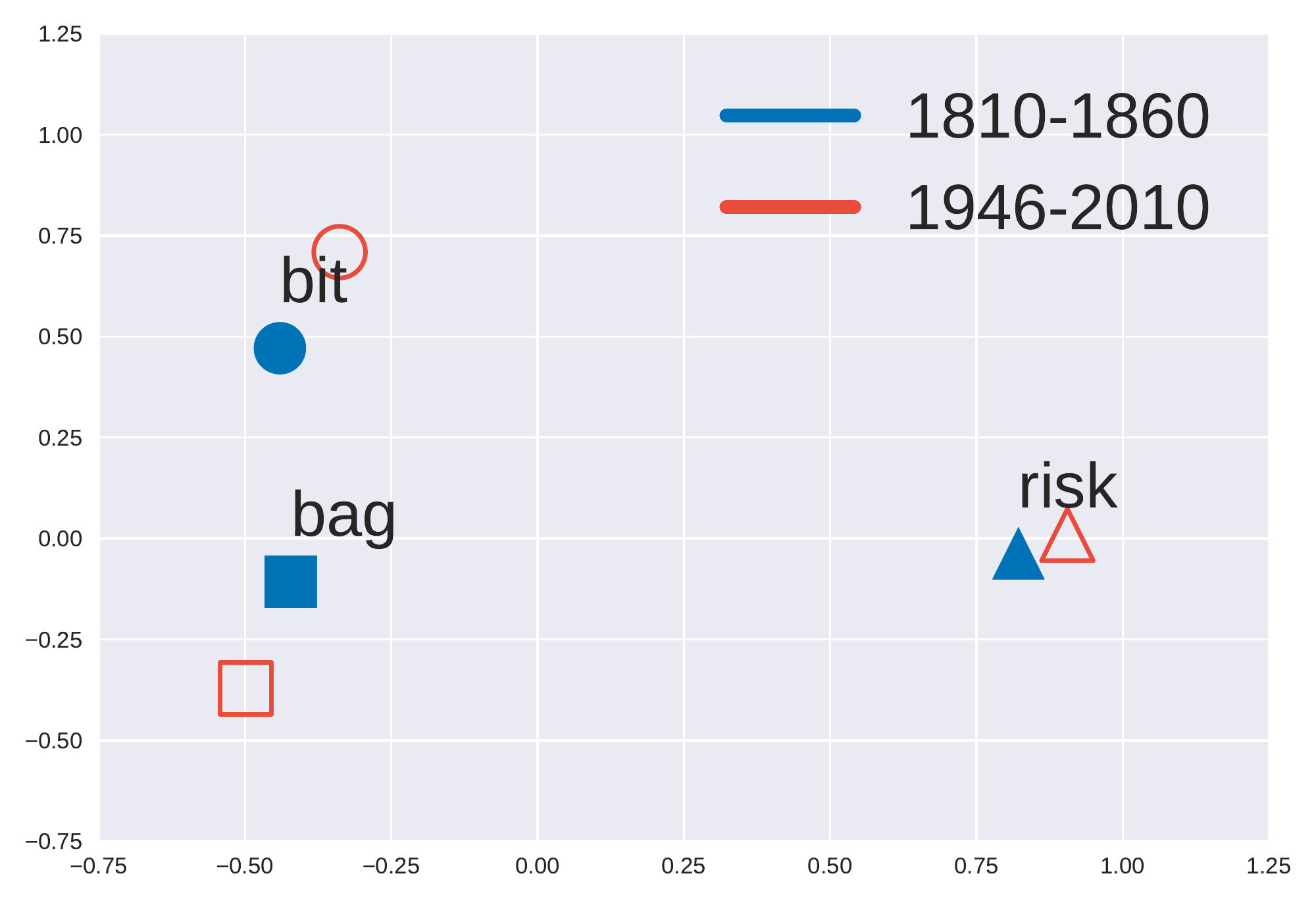}
        \caption{Global Alignment.}
    \end{subfigure}
    \qquad
    \begin{subfigure}{0.4\textwidth}
        \centering
        \includegraphics[width=1\textwidth]{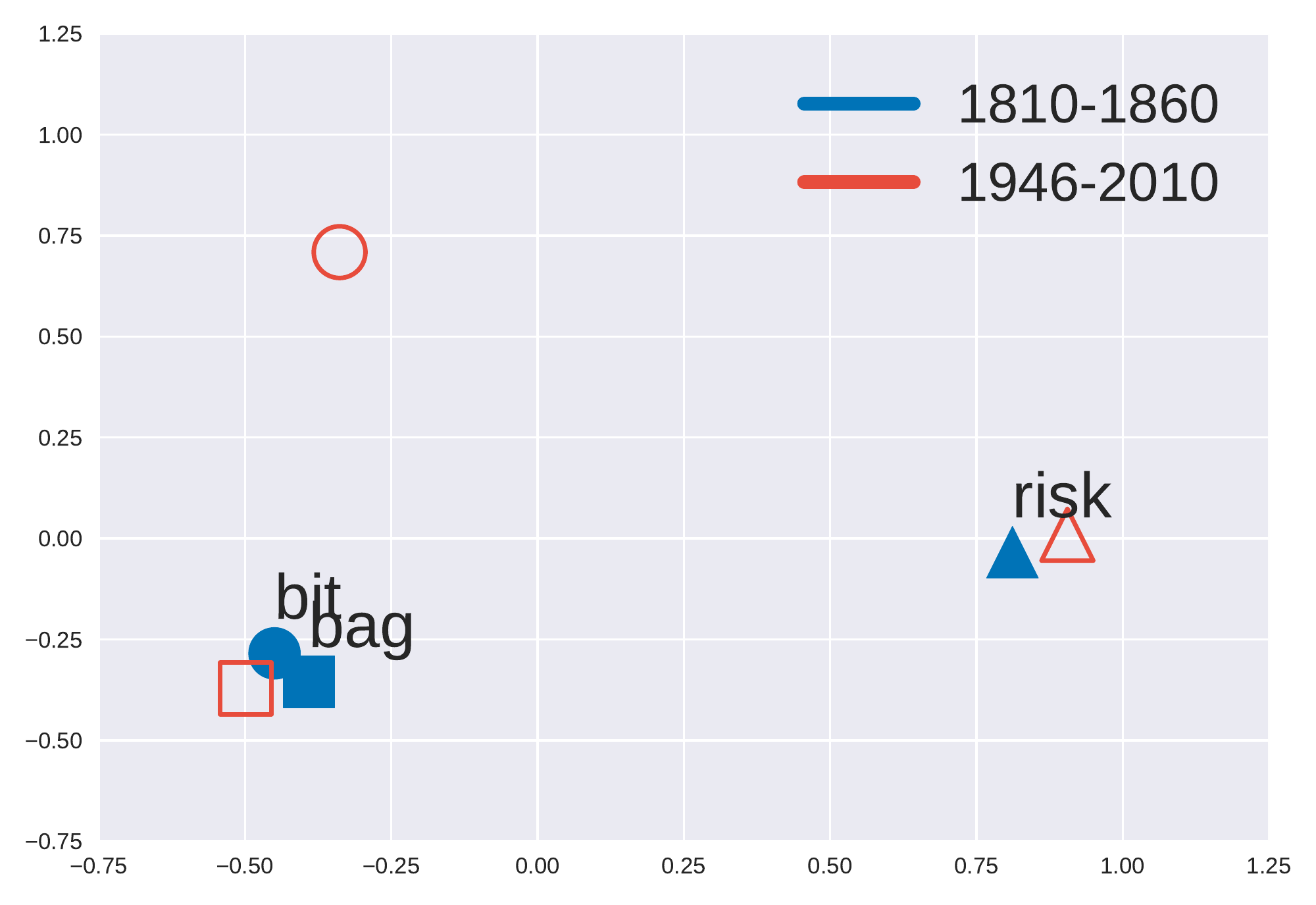}
        \caption{Landmarks selected with \ourmethodA.}
    \end{subfigure}
    \caption{PCA projection of aligned word vectors using: (a) the global alignment strategy, (b) landmarks selected with \ourmethodA. Blue (and red resp.) marks show the location of a word in 1810-1860 (and 1946-2010 resp.). Only the word `bit' is subject to semantic change in this example. Shapes correspond to words \textit{bit} (circle), \textit{bag} (square), \textit{risk} (triangle). Cosine distances of \textit{bit}, \textit{bag}, and \textit{risk} across time periods in are 0.40, 0.41 and 0.01, and 0.61, 0.01, and 0.01 in (b), respectively.
    }
    \label{fig:example}
\end{figure}

This problem is illustrated in Figure \ref{fig:example}. We aligned the embeddings of English historical texts from the 19\textsuperscript{th} and 20\textsuperscript{th}-21\textsuperscript{st} centuries.
The data and ground-truth in this example was extracted from the SemEval-2020 task on unsupervised lexical semantic change detection \cite{schlechtweg2020semeval}. 
The word `bit' is known to have suffered semantic change over time, it has received the new sense of binary digit. 
Words `bag' and `risk' are labelled as semantically stable over the given period. 
We show the results of Global Alignment for this data in  Figure \ref{fig:example}(a) and alignment with landmarks chosen with our proposed method \ourmethodA~in  Figure \ref{fig:example}(b). In Figure \ref{fig:example}(a), by doing Global Alignment, we are forcing 20\textsuperscript{th} century \textit{bit} (red circle) to be close to 19\textsuperscript{th} century \textit{bit} (blue circle) and also 20\textsuperscript{th} century \textit{bag} (red square) to be further away from the 19\textsuperscript{th} century \textit{bag} (blue square). Notice in (b) that by choosing landmarks using \ourmethodA, words \textit{bag} and \textit{risk} from two different time periods remain close, while bit from the different time periods is far, hence amplifying the difference between semantically stable and unstable words. More specifically, the cosine distance across the time periods for \textit{bit} is 0.40, and 0.41 for \textit{bag} in (a). Those respective quantities are 0.61 and 0.01 in (b).

Self-supervised techniques have been shown to be effective in similarly challenging vision-based tasks with limited
(or even w/o) supervision~\cite{dosovitskiy2015discriminative,doersch2015unsupervised,zhang2016colorful,gidaris2018unsupervised}, by using techniques such as image rotation and context changes to create pseudo labels for learning.
In this paper, motivated by these advances, we introduce a new method we call {\em self-supervised semantic shift} (S4 for short) to guide the 
self-supervision by adding perturbations to set of target word vectors by moving them towards other words in the vector space. This process mimics the occurrence of semantic change in the corpora, making specific words semantically unstable. 

This paper makes the following contributions:

\begin{enumerate}
    \item We introduce the novel S4 method and demonstrate its advantage on
    two tasks in monolingual word embedding: Unsupervised Binary Classification, and Word Embedding Alignment. In classification, S4 simulates semantic changes and learns to predict semantically stable and unstable words via self-supervision.
    In alignment, S4 learns to determine the landmark set by predicting, selecting, and refining a set of stable words with self-supervision.
    \item We evaluate \ourmethodD's classification performance on a British v.s. American English semantic change detection task. Regardless of the underlying alignment methods, S4 consistently shows significant gain over the baseline methods, attaining up to 2.65$\times$ higher F1 score.

\item For the SemEval-2020 task \cite{schlechtweg2020semeval}, we show landmark words learned by \ourmethodA attains improved performance on lexical semantic change detection (with a simple linear classifier) in four different languages, demonstrating the importance of landmark selection with \ourmethodA~for  downstream tasks.

\item We also use S4 for discovery of semantic changes in articles within two arXiv subjects: Artificial Intelligence and Classical Physics. We find that S4-based alignment can identify unique semantically changed words in top-ranked word lists that were overlooked by existing alignment methods, providing diversity and novel discovery of lexical semantic changes. 
\end{enumerate}

Our methods have many applications in which semantic change has been shown to be important such as diachronic linguistic analysis \cite{hamilton2016cultural, dubossarsky2017outta}, and predicting armed conflict participants from semantic relations \cite{kutuzov2017temporal}. More recently, Bojanowski et al. \cite{bojanowski2019updating} have presented a strategy using alignment of monolingual embeddings for updating language models to incorporate changes caused by language evolution or usage in specific domains. 

\section{Related Work}


Earliest work on diachronic semantic change discovery analyzes word usage over time with respect to frequency and sense, mostly based on the difference of the distributional property of words between time periods or domains~\cite{sagi2009semantic,cook2010automatically}. 
Distributed word vector representations such as the ones obtained by the skip-gram with negative sampling \cite{mikolov2013distributed} allow for learning distributional information into dense continuous vectors.
Hamilton et al. \cite{hamilton2016diachronic} conducted a diachronic analysis of semantic change on historical text using word embeddings aligned with Orthogonal Procrustes and measuring the displacement of vectors across time periods using cosine distance.
To circumvent the need for alignment, Hamilton et al. \cite{hamilton2016cultural} have proposed a measure of semantic change that compare second-order vectors of distance between a word and its neighbors. Some authors have presented dynamic word embedding techniques to avoid alignment by jointly learning the distributional representations across all time periods \cite{bamler2017dynamic, rudolph2018dynamic, yao2018dynamic}, in which words are connected across time via the assumption that the change across periods is smooth. 
Yin et al. \cite{yin2018global} introduced the Global Anchor method for corpus-level adaptation, which avoids alignment altogether by using second-order distances. This method is proven to be equivalent to the global alignment method and as a result  makes use of the smoothness of change assumption which may lead to oversharing, especially in cross-domain scenarios.

Selection of landmarks as words with likely same meaning in two different languages is used in translation tasks.
Artetxe et al. \cite{artetxe2017learning} employ a self-supervision approach to refine a small seed dictionary with arabic numerals.
Conneau et al. \cite{conneau2017word} use self-supervising adversarial learning to align bilingual embeddings, refining it with Orthogonal Procrustes on the best matching words. 
Joulin et al. \cite{joulin-etal-2018-loss} propose an alternative loss function for the alignment in order to address the conflict between euclidean alignment and cosine distance mapping, and also refines the alignment by matching words that have similar frequency ranking in each language. Lubin et al. \cite{yehezkel-lubin-etal-2019-aligning} employ a landmark selection by detecting noisy pairs through an iterative EM algorithm. 

Most of the previously developed methods do not present a systematic way of detecting semantic change such as in a classification problem. 
Our proposed method is designed explicitly for lexical semantic change tasks by matching the same word in two corpora, approaching semantic change as a binary classification problem. 
Our contributions are: the introduction of a self-supervised method for binary semantic change detection, a method for selecting landmark (anchor) words for alignment of semantically changed word vectors, a quantitative test set on British vs. American English for detecting words with similar and distinct senses. 
We compare our method both to baseline global alignment and noisy pairs methods and show that it provides gains in performance in a number of scenarios.

\section{ Self-Supervised Semantic Shift (S4)}

The problem of detecting semantic change using monolingual alignment of word embeddings is defined as follows. Given two input corpora $\mathcal{A}$ and $\mathcal{B}$, with vocabularies $V_\mathcal{A}$ and $V_\mathcal{B}$, let $A$ and $B$ be the word embedding matrices for the common vocabulary $V = V_\mathcal{A} \cap V_\mathcal{B}$, thus both $A$ and $B$ have dimensions $N \times d$, where $N$ is the size of the common vocabulary, and $d$ is the embedding dimension. 
A word in the common vocabulary is said to be unstable if it is used in a completely different sense between the corpora, or that has multiple senses but some of which only appear in one corpus, other words are considered stable.
One common method for measuring semantic change is to use cosine distance between two embeddings after aligning $A$ and $B$ on a subset of $V$. This problem involves two sub-tasks: detecting words with semantic change, and choosing landmark words to align on. In this paper, we introduce a self-supervised method that can be used in both tasks. 


Given embeddings $A$ and $B$, the main goal of the self-supervision is to create a modified embedding $B'$ such that $B'$ contains a set of words that are known to be semantically shifted with respect to their meaning in $B$ through explicit perturbations. We can generate (pseudo) training samples for a self-supervised procedure by using these modified embeddings and the fact that they are considered semantically shifted.
Suppose $t$ be the target word whose sense we want to add to another word $w$. 
We can accomplish this by replacing $t$ with $w$ an arbitrary number of times $r$ in $\mathcal{B}$ and re-training the word embeddings, where the parameter $r \in (0,1)$ defines the proportion of replacements with respect to the number of occurrences of $t$. To reduce complexity, instead of actually re-training the word embeddings, we move the vector $v_w$ towards $v_t$ by the rule $v_w \gets v_w + rv_t$.
This update rule is derived from the skip-gram with negative sampling model \cite{mikolov2013distributed}, specifically, whenever word $t$ occurs within a neighborhood window of $w$, vector $v_w$ is updated with $v_w + (1-z)\eta v_t$, where $z=\sigma(v_w^\intercal v_t)$. In our perturbation, we replace $\sigma(v_w^\intercal v_t)$ with parameter $r$, which is used to control the proportion in which word $t$ is found in the neighborhood of $w$.
The resulting word vector for $w$ in $\mathcal{B'}$ is now shifted towards that of $t$ and $w$ is now forced to become unstable regardless of its original state.
By applying this process to multiple words, we are able to generate positive samples (semantically changed words) to the self-supervision. In contrast, negative samples are drawn from the set of landmarks.





\begin{algorithm}[ht!]
    \SetAlgoLined
    \KwData{$A, B, L, M, n, r, max\_iters$}
    \KwResult{Classifier weights $W$}
    $W \gets init\_weights$ \;
    $i \gets 0$\;
    
    \While {$i < max\_iters$}{
        $i \gets i + 1$\;
        \tcp{Sample negatives from $L$ and positives from $M$}
        $S_n = uniform\_sample(L, n)$\;
        $S_p = uniform\_sample(M, n)$ \;
        $B' \gets copy(B)$\;
        \For{$w \in S_p$}{
            $t \gets uniform\_sample(M)$\;
            \tcp{Simulate change by moving $w$ towards $t$}
            $B'(w) \gets B(w) + rB(t)$\;
        }
        $ X \gets [A(w), B'(w)]$ $\forall w \in {S_n \cup S_p}$\;
        $ Y \gets [0$ if $w \in S_n$ else $1]$ $\forall w \in {S_n \cup S_p}$\;
        $W \gets train(W, X, Y)$\;
    }
    \Return{$W$}
    \caption{Pseudo-code for self-supervised semantic shift detection (\ourmethodD). Input parameters are word embeddings $A$ and $B$, landmark words set $L$, non-landmark words set $M$, $n$ as the number of negative and positive samples in each iteration, and $r$ is the degree of perturbation. The output of this method is the classifier weights $W$.}    
    \label{alg:selfsup}
\end{algorithm}

\subsection{\ourmethodD: S4 for Semantic Change Detection} \label{sec:s4d}

Self-supervision for semantic change detection can be used in conjunction with any method that uses a subset of words as landmarks.
Given an initial alignment of $A$ to $B$ on a set of landmarks $\mathcal{L}$ (potentially using Orthogonal Procrustes), a batch of positive samples is generated from the perturbations, and negative samples are uniformly drawn from $\mathcal{L}$. These samples are then used to train a binary classifier to predict stable vs. unstable words. We use a single-layer neural network classifier with 100 hidden units with ReLU activation and sigmoid output. The input to the model is the concatenation of row vectors $A(w)$ and $B(w)$ for word $w$. The model is trained over a predefined number of iterations $K$ to predict $\hat{y}=0$ if $w$ is stable, otherwise $\hat{y}=1$. A new batch of positive and negative samples is generated in every iteration. The goal is to minimize the average loss $\frac{1}{K}\sum_{i=1}^K L(\hat{y_i}, y_i)$, with $L$ as the binary cross-entropy function.

Note that, at this point, the self-supervision is done over a fixed alignment of $A$ and $B$ and it is trained to predict semantic change on that setup. The hyper-parameters of \ourmethodD~ are: number of iterations $K$, number of negative and positive samples to generate in each iteration $n$ and $m$, degree of semantic change in the perturbations $r$. The pseudo-code for \ourmethodD~is presented in Algorithm \ref{alg:selfsup}.

\subsection{S4-A: S4 for Alignment of Word Vectors}
\label{sec:s4a}

In this section, we present an extension of the self-supervised training to refine the landmarks based on the classifier predictions from Section~\ref{sec:s4d}, resulting in the Self-Supervised Semantic Shift Alignment (\ourmethodA). 
The general idea is to use stable words for alignment by adding an extra step to each iteration in Algorithm~\ref{alg:selfsup}. At the end of each training iteration, we update the classifiers weights $W$ and use the updated model to predict stable/unstable words across $A$ and $B$, hence updating the set of landmarks $\mathcal{L}$ with words predicted as stable. Finally, we align $A$ to $B$ using the new set of landmarks with orthogonal procrustes and repeat over $K$ iterations. 
This method outputs both model weight and the set of landmark words $\mathcal{L}$ and the set of non-landmark words $\mathcal{N}$. Using the final set $\mathcal{L}$ of landmarks, we can align $A$ to $B$ using orthogonal procrustes on the words in $\mathcal{L}$.  Appendix \ref{app:s4a} contains the pseudo-code for this algorithm.

\section{Experiments}

\subsection{Semantic Change Detection}
\label{sec:ukus}

\noindent \textbf{Objective} We evaluate the ability of \ourmethodD~ to correctly detect the occurrence of lexical semantic change across British and American English. We designed binary classification task to evaluate the model's performance in predicting semantically stable vs. unstable words. The full list of words is given in Appendix \ref{app:ukus}. \\ 

\noindent \textbf{Data set} The corpus used for British English is the British National Corpus (BNC) XML Edition \cite{bnc2007} which contains a mix of news, fiction, and academic texts. 
The corpus for American English is the Corpus of Contemporary American English (COCA) \cite{davies2009385+} which contains text from newspapers, magazines, fiction, and academic texts.
Both corpora are pre-processed by removing stop words, and converting all characters to lower case, resulting in 50M tokens for BNC and 188M tokens for COCA.\\

\noindent \textbf{Baselines} The baseline methods used are the commonly used cosine distance-based method \cite{hamilton2016diachronic, kutuzov-etal-2018-diachronic, schlechtweg-etal-2019-wind}, and the Noise-Aware method \cite{yehezkel-lubin-etal-2019-aligning}, which detects noisy word pairs. 
For the cosine distance (COS), we use three different thresholds for this measure, specifically we have three cosine-based classifiers at thresholds 0.3, 0.5, and 0.7 , above which words are classified as semantically shifted. 
For Noise-Aware, we treat noisy word pairs as semantically shifted (unstable), and clean word pairs as stable, we will refer to that method as \textit{Noisy-Pairs}. 
We compare the baseline methods to our proposed Self-Supervised Semantic Shift Detection (\ourmethodD) with hyper-parameters $n=1000$ positives samples, $m=1000$ negative samples, rate $r=0.25$, trained over 100 iterations (at each iteration, a batch of 1000 positive and 1000 negative samples are generated and used to update the model's weights).
We train Word2Vec on the input corpora after pre-processing with parameters dimension 300, window size 10, minimum count 100 for British, and dimension 300, window size 10, minimum count 200 for American. 
The minimum count for the American corpus is set higher due to its corpus being considerably larger. The common vocabulary contains 26064 words.

\noindent \textbf{Evaluation and Analysis}
Once the embeddings $X_{A}$, $X_{B}$ are learned from the COCA and BNC corpora, respectively, we get the embedding matrices for the common vocabulary $A, B \in \mathbb{R}^{n \times d}$ by selecting the rows of $X_{A}$ and $X_{B}$ that correspond to words in the common vocabulary. 
We learn a transform matrix $Q$ by aligning $A$ to $B$ using a given alignment strategy.
The self-supervised classifier is trained on $A$ and $B$ using the given alignment.
Using the learned matrix $Q$, we align $X_{B}$ to $X_{A}$ by doing $X_{B} \gets X_{B}Q$.
Finally, we concatenate the vectors of the target words and feed it into the classifier to obtain the binary predictions (i.e. semantically stable or unstable).

The classification scores for this task (Table \ref{tab:ukus}) show that \ourmethodD~ displays the best accuracy when aligning on the top 10\% most frequent words, it also shows high recall and F1 scores when aligning on the 10\% least frequent and 5\% most frequent words. 
The scores for \ourmethodD~ are the average of 10 evaluation rounds, standard deviation is also reported in Table \ref{tab:ukus}. 
Each evaluation round consists of one execution of the algorithm on the input data.
This contrasts with the drop in performance shown by global alignment, this is likely due to the oversharing of words, which makes the separation of stable and unstable words more difficult.
The alignment method is irrelevant to Noisy-Pairs since it inherently aligns the input vectors when searching for noise. 
Noisy-Pairs predicts a total of 24659 pairs as clean, or semantically stable. For that reason, only one pair of words from the target set is predicted as semantically shifted (positive class), which explains the precision score of 1.0 and a recall of 0.03. 

Examples of stable words correctly predicted by \ourmethodD~ are the British-American pairs \textit{labour/labor}, \textit{defence/defense}, \textit{petrol/gas}, \textit{football/soccer}, and \textit{queue/line}. 
This shows we are able to not only to detect identical words but also morphological differences and synonyms.
Note that some of these words were not included in the alignment due to not being in the common vocabulary. Yet, we are still able to capture their semantic similarity after the orthogonal transformation. These results show our model's ability to generalize to words not seen in the self-supervision.
Additionally, we were able to correctly predict unstable words such as \textit{chips} (french fries in the US), \textit{biscuit} (scone in the UK, cookie in the US), and \textit{semi} (house in the UK, truck in the US). Noisy-Pairs is able to correctly predict the semantically unstable words \textit{subway} and \textit{yankee}, and it also predicts all stable words correctly but shows a low recall score.

\begin{table*}[ht]
    \centering
    \caption{Classification scores for the British vs. American English task. The baseline cosine (COS) method outputs unstable words whose cosine distance is greater than $0.3/0.5/0.7$. Top-N/bot-N alignments use the most/least frequent words. \ourmethodD~ (our method) is the self-supervised semantic shift detection. The scores for \ourmethodD~ are given as the mean and standard deviation over 10 evaluation rounds. The initial alignment is irrelevant to noisy-pairs as it necessarily searches an alignment.
    }
    \begin{tabular}{llrrrr}
         \textbf{Method}                & \textbf{Alignment} & \textbf{Accuracy} & \textbf{Precision} & \textbf{Recall} & \textbf{F1} \\ \toprule
         COS                     & \multirow{2}{*}{Global}  & $0.38/0.30/0.25$ & $0.48/0.36/0.00$ & $0.26/0.10/0.00$ & $0.32/0.15/0.00$ \\
         \ourmethodD                 &  & $0.45$ $\pm 0.03$ & 0.81 $\pm 0.06$ & $0.28 \pm 0.05$ & 0.43 $\pm 0.06$ \\   \midrule    
         COS                        & \multirow{2}{*}{Top-5\%} & $0.39/0.30/0.25$ & $0.47/0.36/0.00$ & $0.29/0.10/0.00$ & $0.33/0.15/0.00$ \\
         \ourmethodD                 &  & 0.67 $\pm 0.02$ & $0.81 \pm 0.02$ & $0.79 \pm 0.02$ & \textbf{0.82} $\pm 0.01$ \\ \midrule
         COS                        & \multirow{2}{*}{Top-10\%} & $0.37/0.30/0.25$ & $0.47/0.36/0.00$ & $0.26/0.10/0.00$ & $0.31/0.15/0.00$ \\
         \ourmethodD                 &  & $\mathbf{0.70} \pm 0.03$ & $\mathbf{0.83} \pm 0.02$ & $0.74 \pm 0.03$ & $0.78 \pm 0.02$\\ \midrule
         COS                        & \multirow{2}{*}{Bot-5\%} & $0.40/0.29/0.24$ & $0.46/0.32/0.00$ & $0.31/0.10/0.00$ & $0.35/0.16/0.00$ \\
         \ourmethodD                 & & $0.45 \pm 0.02$ & $0.62 \pm 0.01$ & $0.21 \pm 0.02$ & $0.31 \pm 0.01$ \\ \midrule
         COS                        & \multirow{2}{*}{Bot-10\%} & $0.37/0.30/0.25$ & $0.53/0.45/0.25$ & $0.29/0.12/0.02$ & $0.34/0.18/0.03$ \\
         \ourmethodD                 &  & $0.65 \pm 0.02$ & $0.71 \pm 0.01$ & $0.89 \pm 0.02$ & $0.79 \pm 0.01$ \\ \midrule
         COS        & \ourmethodA       & $0.44/0.34/0.28$  & $0.66/0.62/0.57$ & $0.40/0.18/0.06$  & $0.44/0.27/0.12$ \\ 
         \ourmethodD    & \ourmethodA   & $\mathbf{0.70} \pm 0.01$  & $0.72 \pm 0.01$  & $\mathbf{0.93} \pm 0.01$   & $ 0.81 \pm 0.01 $\\ \midrule
         Noisy-Pairs                & -  & $0.30$ & $\mathbf{1.00}$ & $0.03$ & $0.06$ \\ \bottomrule

    \end{tabular}
    \label{tab:ukus}
\end{table*}

\subsection{Evaluating Alignment Strategies}
\label{sec:semeval}

\noindent \textbf{Objective}
We evaluate the Self-Supervised Semantic Shift Alignment (\ourmethodA) described in Section \ref{sec:s4a} by the impact of assessing multiple alignment strategies on the performance of the binary classification problem from the recent SemEval-2020 task on Unsupervised Lexical Semantic Change Detection \cite{schlechtweg2020semeval}. The task consists of predicting the occurrence of lexical semantic change on a set of target words in four languages: English, German, Latin, and Swedish. For each language there are two input corpora $C_1$ and $C_2$ containing text from time periods $t_1$ and $t_2$, with $t_1 < t_2$. 

\noindent \textbf{Data sets}
The data sets used in this experiment are provided in the aforementioned SemEval task. The corpus used for English is the Clean Corpus of Historical American English (CCOHA) \cite{alatrash2020clean}, a pre-processed and lemmatized version of the Corpus of Historical American English (COHA) \cite{davies2009385+}. For this task, the corpus was split into time periods $t_1$ with texts from years 1810 through 1860, and $t_2$ with text from years 1960 through 2010.
The German data set consists of the DTA corpus \cite{dta2018corpus} for the first time period, and a combination of the BZ and ND corpora \cite{bz2018corpus, nd2018corpus} for the second time period. Specifically, text in $t_1$ pertains to years 1800 through 1899, and text in $t_2$ pertains to years 1946 through 1990.
For Latin we use the LatinISE corpus \cite{mcgillivray2013tools} with time periods $t_1$ from 200 B.C. through 0 A.D., and $t_2$ from years 0 through 2000 (A.D.).
For Swedish we use the KubHist corpus \cite{borin-etal-2012-korp, adesam2019exploring}. Time period $t_1$ is from 1790 through 1830, time period $t_2$ is from 1895 through 1903. Along with each data set, a list of target words is provided for evaluation. Further details about the data sets can be found in Appendix \ref{app:semeval}.

\noindent \textbf{Baselines}
We include common alignment methods for comparison to \ourmethodA. Particularly, we adopt the global alignment strategy as used in diachronic embeddings \cite{hamilton2016diachronic}, as well as frequency based selection of landmarks, aligning at the 5\% and 10\% most and least frequent words \cite{bojanowski2017enriching}, and the Noise-Aware method for selecting landmarks.

\noindent \textbf{Evaluation and Analysis} We begin by training Word2Vec on $C_1$ and $C_2$, generating embeddings $X_1$ and $X_2$. We set the embedding dimension to 300 and use a window of size 10 for all languages. The minimum word count is 20, 30, 10, and 50 for English, German, Latin, and Swedish, respectively, these are chosen based on the amount of data provided for each language.

Let $A \subset X_1$ and $B \subset X_2$ be the embedding matrices for the common vocabulary terms in $C_1$ and $C_2$. Our experiment consists of aligning $A$ to $B$ using different alignment strategies and evaluating the alignments with respect to its performance in the binary classification task. Particularly, we evaluate our self-supervised alignment and compare it to doing global alignment, aligning on most and least frequent words, and selecting clean words as landmarks \cite{yehezkel-lubin-etal-2019-aligning}.

Since there is no labeled training 
data  for this problem, we build a model to predict lexical semantic change based on the cosine distance between the word vectors in $A$ and $B$, after alignment. 
We compute the cosine distance between all pairs of vectors in $A$ and $B$. Then, for each word $w$ in the common vocabulary, we compute the cumulative probability $P(x<X)$ where $x$ is the cosine distance between the vectors of $w$. 
Finally, we decide on the class of $w$ based on a threshold $t \in (0, 1)$.
To determine the value of $t$, we perform model selection through cross-validation on the self-supervised data, selecting $t$ that achieves the best accuracy in the leave-one-out tests, $t$ is searched in $(0, 1)$ in increments of $0.1$.
The prediction is $\hat{y}=1$ if $P(x<X) > t$, otherwise $\hat{y}=0$.

Results from this experiment are shown in Table \ref{tab:semeval}. 
We report the accuracy on the evaluation set for each language and alignment method. 
Top/bot alignments are done over the top 5\%/10\% most/least frequent words in $C_1$. 
\ourmethodA~ is able to achieve the best accuracy scores maximum scores for English and German, matching some of the top performing scores in the post-evaluation phase of the SemEval-2020 Task 1 competition \footnote{\url{https://competitions.codalab.org/competitions/20948#results}}.

\begin{table}[ht]
    \centering
    \caption{Classification accuracy of the unsupervised lexical semantic change detection on the SemEval-2020 Task 1 data set. 
    The results were obtained by aligning the embedding matrices using different alignment strategies, and applying a threshold to the cosine distance of the aligned vectors, selected by cross-validation. 
    Top-fr. and Bot-fr. are alignments using OP on the top and bottom $5\%$ and $10\%$ frequent words. 
    The bottom rows shows the top 3 high scoring submissions to SemEval-2020 Task 1 in the post evaluation phase.}
    \begin{tabular}{lrrrr}
         \textbf{Alignment} & \multicolumn{1}{c}{\textbf{English}} & \multicolumn{1}{c}{\textbf{German}} & \multicolumn{1}{c}{\textbf{Latin}} & \multicolumn{1}{c}{\textbf{Swedish}}  \\ \toprule

         \ourmethodA     & \textbf{0.70} & \textbf{0.81} & 0.68 & 0.77 \\
         Noise-Aware    & 0.65 & 0.79 & 0.65 & 0.74 \\
         Top 5\% fr.    & 0.65 & 0.77 & 0.68 & 0.77 \\
         Top 10\% fr.   & 0.68 & 0.79 & 0.68 & 0.74 \\
         Bot 5\% fr.    & 0.68 & 0.73 & 0.62 & 0.77 \\
         Bot 10\% fr.   & 0.68 & 0.75 & \textbf{0.70} & \textbf{0.81} \\
         Global         & 0.68 & 0.79 & 0.65 & 0.74 \\ \midrule
        SemEval \#1 & \multicolumn{1}{r}{\textbf{0.70}} & \multicolumn{1}{r}{0.79} & \multicolumn{1}{r}{0.68} & \multicolumn{1}{r}{\textbf{0.81}} \\
        SemEval \#2 & \multicolumn{1}{r}{\textbf{0.70}} & \multicolumn{1}{r}{0.77} & \multicolumn{1}{r}{0.72} & \multicolumn{1}{r}{0.74} \\
        SemEval \#3 & \multicolumn{1}{r}{0.65} & \multicolumn{1}{r}{0.77} & \multicolumn{1}{r}{0.75} & \multicolumn{1}{r}{0.74} \\
        \bottomrule
         
    \end{tabular}
    \label{tab:semeval}
\end{table}

\subsection{Discovery of Semantic Change}
\label{sec:discovery}

\noindent \textbf{Objective}
We conduct an experiment on the arXiv data provided by Yin et al. \cite{yin2018global} to show how we can use \ourmethodA~ for word embedding alignment for the discovery of semantic change, and how the results  differ across alignment methods. 
We select the subjects of Artificial Intelligence (cs.AI) and Classical Physics (physics.class-ph) and train embeddings $A$ and $B$, respectively, with Word2Vec (dimension 300, window size 10, minimum count 20).
The embedding matrices are aligned using each alignment strategy, and the semantic shift measured by $d_i = \left\lVert A_iQ - B_i \right\rVert$ for each word $w_i$ in the common vocabulary, where $Q$ is the transform matrix learned in the alignment.

\noindent \textbf{Baselines}
We compare the most semantically shifted words as discovered by the Global and Noise-Aware alignments \cite{hamilton2016diachronic, yehezkel-lubin-etal-2019-aligning}. We also compare our results to the top 3 high scoring entries from post-evaluation phase of the SemEval-2020 Task 1 competition, these methods may use distinct sets of features that go beyond just using word embeddings.

\begin{figure}[ht]
    \begin{minipage}{0.45\textwidth}
        \centering
        \includegraphics[width=0.9\textwidth]{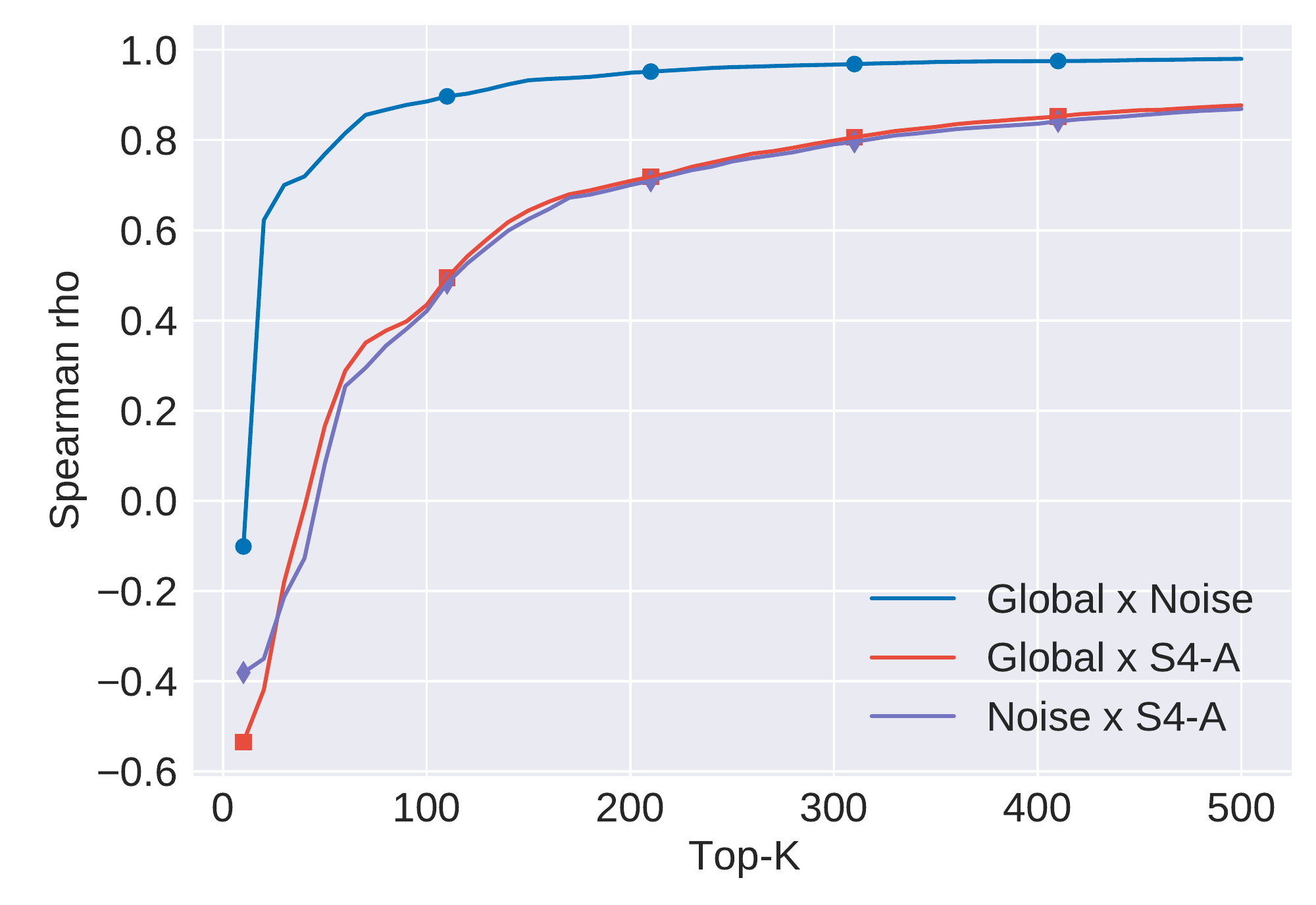}
        \caption{Ranking correlation between Global, Noise-Aware, and \ourmethodA~ alignments at varying top-k levels. Works are ranked from most to least shifted according to each method and the ranking correlation is measured with Spearman's rho. The semantic shifts are between arXiv subjects cs.AI and physics.class-ph.}
        \label{fig:ranking}
    \end{minipage}
    ~
    \begin{minipage}{0.45\textwidth}
    \centering
    \captionsetup{type=table}
    \caption{Unique words discovered by each alignment method among the top 50 most shifted, between arXiv cs.AI and physics-class-ph corpora. Global and Noise-Aware show the same predictions in the top 50 words.}
    \begin{tabular}{cc}
        \textbf{Global/Noise-Aware} & \textbf{\ourmethodA} \\ \toprule
        agent           & component \\
        approximation   & element  \\
        boundary        & mass \\
        conceptual      & order \\
        knowledge       & solution \\
        plane           & space \\
        reference       & term \\
        rules           & time \\
        system          & vector \\
     \bottomrule
    \end{tabular}
    \label{tab:words}
    \end{minipage}
\end{figure}

\noindent \textbf{Evaluation and Analysis}
To quantify the difference between different alignments, we measured the ranking correlation using the Spearman's rho coefficient of the ranked list of words according to each method (ranked in descending order of semantic shift) at  varying top-K thresholds with $k$ in $[10, 500]$ in increments of $10$. 
Figure \ref{fig:ranking} shows the ranked correlation coefficient between each alignment strategy. Higher values of rho indicate that the order of semantic shift is more consistent between the two alignment strategies.
These results reveal that Global and Noise-Aware produce very similar rankings, with rho approaching $1$ even for small values of $k$. On the other hand, the ranking correlation between \ourmethodA~ is substantially lower for small values of $k$.
This suggests that most of the difference in ranking between \ourmethodA~ and the others is in the most shifted words, with the ranking of the remaining words being very similar to Global and Noise-Aware.
In summary, \ourmethodA~ can be used to find novel shifted words that are overlooked by existing methods such as Global and Noise-Aware. Table \ref{tab:words} shows the list of uniquely discovered words among the top most shifted for Global and \ourmethodA~. Noise-Aware because it does not show any novel words when compared to Global, i.e., its predictions are the same between arXiv subjects of Artificial Intelligence and Classical Physics. We find that words uniquely discovered by \ourmethodA~ can be naturally explained in context of their subjects, for instance, \textit{mass} is likely more often used as probability mass in AI, and as physical mass in classical physics. More comparisons are shown in Appendix \ref{app:arxiv}.

\section{Conclusions}

We introduced \ourmethodD~and \ourmethodA~as self-supervised approaches to detect word-level semantic shifts on monolingual corpora. 
Motivated by the unsupervised nature of this problem, we introduce self-supervision based on the perturbation of word vectors and apply it to binary classification and vector alignment. \ourmethodD~ is presented as an alternative to baseline unsupervised methods for semantic shift detection, particularly in the case of binary classification. 
We show, through experiments in Section \ref{sec:ukus}, that it achieves over 2$\times$ higher F1-scores than baselines in the classification settings.
Moreover, we show how the alignment of word embeddings affect the outcome of such methods. 
Particularly, we show that global alignment uses the assumption of smooth transition, which may not hold true in the scenario of cross-domain semantic shift, where many words can be highly shifted. 
For that reason, we present an extension of our method, named \ourmethodA, that uses its predictions to refine the alignment of the input embeddings. 
We demonstrate its usefulness quantitatively, through the detection task in Section \ref{sec:semeval}, where \ourmethodA~allows for the detection of unique words when using a simple cosine distance baseline. 
Qualitatively, we demonstrate that \ourmethodA~is able to discovery novel shifts when compared to other alignment methods.

There are still open questions on how the self-supervised model is affected according to part-of-speech, frequency range, and degree of polysemy of words. In addition, factors such as number of tokens, vocabulary size, and degree of change of the language may impact the quality of the embeddings, therefore, affect the semantic shift detection.

While this remains a difficult task, we believe that this work will help numerous applications of semantic shift detection and alignment that have been recently explored, especially in the monolingual and cross-domain setting.

\section*{Ethical Impact}
Prior work has shown that embeddings in a single corpus can encode many cultural stereotypes such as gender and racial bias. This is not particularly surprising as language is a tool for creating common meaning. Stereotypes, gender and racial bias are social constructs that have been quite prominent in many cultural artifacts including languages. Code words and dog-whistles are words often used to mean different things only for a specific community. A word that looks quite common and benign can be very offensive when evaluated in context. Some of these differences can be rooted in social justice work of reclaiming controversial words. In short, language codes rich and complex cultural and historical differences that are lost when treated as a monolith. Our emphasis is in creating tools not to smooth over these differences, but help identify them. 
To accomplish this, we use two corpora in the same language to understand which words in one corpus have a different meaning compared to another corpus. 
Our work leverages potential bias in languages by using one corpus as a point of reference for the other to highlight the differences.

\section*{Acknowledgements}
This work was supported by the Rensselaer-IBM AI Research Collaboration (\url{http://airc.rpi.edu}), part of the IBM AI Horizons Network (\url{http://ibm.biz/AIHorizons}).
We also thank the anonymous reviewers who contributed with our work with their invaluable feedback.

\bibliography{main}

\clearpage
\appendix

\section{Self-Supervised Semantic Shifts}
\subsection{Pseudo-code for \ourmethodA~ (Section \ref{sec:s4a})}
\label{app:s4a}

Algorithm \ref{alg:s4a} shows the pseudo-code for Self-Supervised Semantic Shift Alignment from Section \ref{sec:s4a}.

\begin{algorithm}[h!]
    \SetAlgoLined
    \KwData{$A, B, L, M, n, r, max\_iters$}
    \KwResult{List of landmark words $L$, and non-landmarks $M$}
    $W \gets init\_weights$ \;
    $i \gets 0$\;
    
    \While {$i < max\_iters$}{
        $i \gets i + 1$\;
        $A \gets Align(A, B, L)$ \tcp*{Align with OP}
        \tcp{Sample negatives from $L$ and positives from $M$}
        $S_n = uniform\_sample(L, n)$\;
        $S_p = uniform\_sample(M, n)$ \;
        $B' \gets copy(B)$\;
        \For{$w \in S_p$}{
            $t \gets uniform\_sample(M)$\;
            \tcp{Simulate change by moving $w$ towards $t$}
            $B'(w) \gets B(w) + rB(t)$\;
        }
        $ X \gets [A(w), B'(w)]$ $\forall w \in {S_n \cup S_p}$\;
        $ Y \gets [0$ if $w \in S_n$ else $1]$ $\forall w \in {S_n \cup S_p}$\;
        $W \gets train(W, X, Y)$\;
        $\hat{Y} \gets predict(words, A, B, W)$\;
        $L \gets [w]$ $ \forall w \in words$ if $\hat{Y} = 0$ \tcp{Update landmark set}
        $M \gets [w]$ $ \forall w \in words$ if $\hat{Y} = 1$\;
    }
    \Return{$L, M$}

    \caption{Pseudo-code of \ourmethodA. Input parameters are word embeddings $A$ and $B$, list $words$, $n$ is the number of negative and positive samples in each iteration, $r$ is the rate of semantic change.}
    \label{alg:s4a}
\end{algorithm}

\section{British-American English Experiment (Section \ref{sec:ukus})}
\label{app:ukus}

\paragraph{Pre-processing} 
Both corpora were pre-processed by removing stop words in NLTK's list of English stop words \footnote{\url{https://www.nltk.org/}}, tokenizing by splitting at non-word characters (spaces, punctuation, symbols), and converting characters to lower case.

\paragraph{British National Corpus (BNC)} 
Text Categories: fiction, newspaper, academic. Tokens: 50M (after pre-processing). Vocabulary size: 45813 words. Publication dates: 1960-1993.

\paragraph{Corpus of Contemporary American English (COCA)} 
Text Categories: fiction, newspaper, magazine, academic. Tokens: 188M (after pre-processing). Vocabulary size: 30415 words. Publication dates: 1990-2012.

\paragraph{Word2Vec Parameters}
Word embeddings were trained using Gensim's implementation of Word2Vec skip-gram with negative sampling \footnote{\url{https://radimrehurek.com/gensim/index.html}}. 
The parameters are described below, for each corpus.

\paragraph{\ourmethodD~ Parameters}
Number of positive samples $n$=1000. Number of negative samples $m=1000$. Degree of perturbation $r=0.25$. Iterations: 100.

\begin{table}[ht!]
    \centering
    \caption{Word2Vec parameters used to train embeddings on BNC and COCA.}
    \begin{tabular}{lcc}
        \textbf{Parameter}&   \textbf{BNC} & \textbf{COCA} \\ \toprule
        Dimensionality & 100 & 100   \\
        Window size & 10 & 10  \\
        Min. word count & 100 & 200  \\
        Negative samples& 5 & 5  \\
    \end{tabular}
    \label{tab:w2v_ukus}
\end{table}

\subsection{List of words in the evaluation set}
List of words used in British-American English and their respective labels in section \ref{sec:ukus}. The list of stable words is from American English at United States Department of State \footnote{\url{http://americanenglish.state.gov/}}. We only use unigrams from the word lists, n-grams are ignored 
\footnote{Stable words: https://designyoutrust.com/2015/11/british-american-english-differences/}
\footnote{Unstable words: https://www.usingenglish.com/articles/100-words-with-different-meanings-in-british-american-english.html}. 
The final evaluation set consists of 67 stable and 78 unstable word pairs. Table \ref{tab:ukus_similar} shows the list of stable word pairs, i.e., the pairs that belong to class 0. Table \ref{tab:ukus_dissimilar} shows the list of word unstable words. In the unstable case, the pairs are formed by identical words that have different meanings, which belong to class 1. E.g.: we compare the word vectors of word \textit{subway} in the British and American embeddings.

\begin{table*}[ht!]
    \centering
    \caption{List of British and American English word pairs that are labeled as semantically similar (stable). These words include synonyms and words that differ in spelling and should be classified as semantically stable (class 0).}
    \begin{tabular}{cc|cc}
        \textbf{British English} & \textbf{American English} &
        \textbf{British English} & \textbf{American English} \\ \hline
        aeroplane & airplane & manoeuvre & maneuver \\
        analogue & analog & mobile & cellphone  \\
        analyse & analyze  & motorway & highway\\
        apologise & apologize & nappy & diaper \\
        aubergine & eggplant & paediatric & pediatric\\
        autumn & fall & paralyse & paralyze\\
        behaviour & behavior & pavement & sidewalk \\
        bonnet & hood & petrol & gas \\
        boot & trunk &  pitch & field \\
        braces & suspenders & post & mail \\
        breathalyse & breathalyze & pretence & pretense \\
        catalogue & catalog & pullover & sweater \\
        cheque & check & queue & line \\
        colour & color & recognise & recognize\\
        courgette & zucchini & rubber & eraser \\
        defence & defense & rubbish & garbage\\
        dialogue & dialog & rumour & rumor \\
        draw & tie & sweets & candy \\
        dummy & pacifier & takeaway & takeout  \\
        favourite & favorite & telly & tv\\
        flat & apartment & theatre & theater\\
        flavour & flavor & timetable & schedule\\
        football & soccer & torch & flashlight\\
        grey & gray & trainers & sneakers\\
        humour & humor & travel & travel\\
        kit & uniform & travelled & traveled \\
        labour & labor & traveller & traveler\\
        leukaemia & leukemia & travelling & traveling\\
        licence & license & treacle & molasses\\
        lift & elevator & underground & subway\\
        loo & restroom & waistcoat & vest\\
        lorry & truck \\
        
    \end{tabular}
    \label{tab:ukus_similar}
\end{table*}

\begin{table*}[ht!]
    \centering
    \caption{List of words labeled as semantically different (unstable) across British and American English. These words are used in different senses between British and American English and should be classified as unstable (class 1). Note that we use the same word on both corpora for comparison, unlike in the stable case. E.g.: we compare word vectors of word \textit{subway} in British and American English.}
    \begin{tabular}{cccc}
    \multicolumn{4}{c}{\textbf{Unstable words}} \\ \toprule
    appropriation & dc & mortuary & sherbet \\
    asian & dormitory & nappy & silverware \\
    athlete & entree & nervy & sprouts \\
    bathroom & faculty & pantomime & squash \\
    bill & flapjack & pants & strike \\
    biscuit & football & parentheses & subway \\
    brackets & gas & pavement & surgery \\
    bum & grill & penny & suspenders \\
    buzzard & gym & pissed & sweets \\
    campsite & hamper & professor & tights \\
    casualty & hockey & prom & tosser \\
    chips & homely & pudding & trailer \\
    cider & hood & purse & trolley \\
    commissioner & hooker & rider & trooper \\
    commonwealth & jelly & robin & tuition \\
    constable & jock & roommate & underpass \\
    cooker & jumper & rubber & vest \\
    corn & mad & saloon & yankee \\
    cracker & medic & semi \\

    \end{tabular}
    \label{tab:ukus_dissimilar}
\end{table*}

\section{SemEval Experiment (Section \ref{sec:semeval}) }
\label{app:semeval}

In this section we describe the corpora utilized for each language in the experiment from Section \ref{sec:semeval}. Each uses two corpora pertaining to two different time periods $t_1$ and $t_2$. Time periods are shown in Table \ref{tab:semeval_stats}.

\paragraph{Pre-processing}
Pre-processing was done by tokenizing and converting characters to lower case. Stop word removal was applied to English, German, and Swedish corpora.

\paragraph{Word2Vec Paramters} Word2Vec parameters are shown in Table \ref{tab:w2v_semeval}.

\paragraph{\ourmethodA~ Parameters} $n$ - number of positive samples; $m$ - number of negative samples; $r$ - degree of perturbation; $iters$ - number of iterations.
\begin{itemize}
    \item \textbf{English}: $n=100$, $m=50$, $r=1$, $iters=100$.
    \item \textbf{German}: $n=100$, $m=200$, $r=1$, $iters=100$.
    \item \textbf{Latin}: $n=10$, $m=4$, $r=0.5$, $iters=100$.
    \item \textbf{Swedish}: $n=100$, $m=200$, $r=1$, $iters=100$.
\end{itemize}

\begin{table*}[ht!]
    \centering
    \caption{Corpora used in SemEval-2020 Task 1 for each language, with their respective time periods, and size of common vocabulary.}
    \begin{tabular}{llccc}
        \textbf{Language} & \textbf{Corpora} & $t_1$ & $t_2$ & \textbf{Size of common vocab.} \\ \toprule
        English & CCOHA & 1810-1860 & 1960-2010 & 8079 \\
        German & DTA, BNZ, ND & 1800-1899 & 1946-1990 & 25905 \\
        Latin & LatinISE & -200-0 & 0-2000 & 7689 \\
        Swedish & KubHist & 1790-1830 & 1895-1903 & 20748
    \end{tabular}
    \label{tab:semeval_stats}
\end{table*}

\begin{table*}[ht!]
    \centering
    \caption{Word2Vec parameters used to train embeddings on SemEval data sets.}
    \begin{tabular}{lcccc}
        \textbf{Parameter}&   \textbf{English} & \textbf{German} & \textbf{Latin} & \textbf{Swedish} \\ \toprule
        Dimensionality & 300 & 300 & 300 & 300   \\
        Window size & 10 & 10 & 20 & 10 \\
        Min. word count & 20 & 30 & 10 & 50 \\
        Negative samples& 5 & 5 & 5 & 5 \\
    \end{tabular}
    \label{tab:w2v_semeval}
\end{table*}

\subsection{Target Words}

The list of labeled target words from the SemEval task is shown in Table \ref{tab:semeval_targets}.

\begin{table*}[ht]
    \centering
    \caption{List of target words for the SemEval task. Label 0 means the word is semantically stable, 1 indicates semantically unstable words. The provided English corpus is lemmatized, lemmas are shown by the \_ suffix.}
    \begin{tabular}{lcp{0.7\textwidth}}
    
    \multirow{8}{*}{\textbf{English}} & \multirow{4}{*}{Stable} & bag\_nn, ball\_nn, chairman\_nn, contemplation\_nn, donkey\_nn, face\_nn, fiction\_nn, gas\_nn, lane\_nn, multitude\_nn, ounce\_nn, part\_nn, pin\_vb, quilt\_nn, relationship\_nn, risk\_nn, savage\_nn, stroke\_vb, tree\_nn, twist\_nn, word\_nn \\ \\
                                      & \multirow{3}{*}{Unstable} & attack\_nn, bit\_nn, circle\_vb, edge\_nn, graft\_nn, head\_nn, land\_nn, lass\_nn, plane\_nn, player\_nn, prop\_nn, rag\_nn, record\_nn, stab\_nn, thump\_nn, tip\_vb \\ \midrule
    \multirow{10}{*}{\textbf{German}} & \multirow{6}{*}{Stable} & abgebrüht, Ackergerät, Armenhaus, aufrechterhalten, Ausnahmegesetz, beimischen, Einreichung, Eintagsfliege, Entscheidung, Festspiel, Frechheit, Fuß, Gesichtsausdruck, Kubikmeter, Lyzeum, Mulatte, Naturschönheit, Pachtzins, Seminar, Spielball, Tier, Titel, Tragfähigkeit, Truppenteil, Unentschlossenheit, vergönnen, voranstellen, vorliegen, vorweisen, weitgreifend, zersetzen\\ \\
                                    & \multirow{3}{*}{Unstable} & abbauen, abdecken, Abgesang, artikulieren, ausspannen, Dynamik, Engpaß, Knotenpunkt, Manschette, Mißklang, Ohrwurm, packen, Rezeption, Schmiere, Sensation, überspannen, verbauen\\ \midrule
    \multirow{6}{*}{\textbf{Latin}} & \multirow{2}{*}{Stable} & acerbus, ancilla, consilium, fidelis, honor, hostis, itero, necessarius, nobilitas, oportet, poena, salus, sapientia, simplex \\ \\
                                    & \multirow{3}{*}{Unstable} & adsumo, beatus, civitas, cohors, consul, credo, dolus, dubius, dux, humanitas, imperator, jus, licet, nepos, pontifex, potestas, regnum, sacramentum, sanctus, scriptura, senatus, sensus, templum, titulus, virtus, voluntas \\ \midrule
    \multirow{6}{*}{\textbf{Swedish}} & \multirow{3}{*}{Stable} & aktiv, annandag, antyda, bearbeta, bedömande, beredning, blockera, bolagsstämma, bröllop, by, central, färg, förhandling, gagn, kemisk, kokärt, notis, studie, undertrycka, uppfostran, uträtta, vaktmästare, vegetation\\ \\
                                        & \multirow{2}{*}{Unstable} & granskare, konduktör, krita, ledning, medium, motiv, uppfattning, uppläggning \\ 

    \end{tabular}
    \label{tab:semeval_targets}
\end{table*}

\newpage
\subsection{Landmark Sets Learned by \ourmethodA~ Are Stable}

Results show that \ourmethodA~ converges to a stable set of landmarks over time. We demonstrate that by applying \ourmethodA~ to the SemEval data sets, computing the overlap of the landmarks $\mathcal{L}$ of consecutive iterations measured by the Jaccard Index. Specifically, for iteration $i$, we compute:

$$J(i, i-1) = \frac{|\mathcal{L}_{i} \cap \mathcal{L}_{i-1}|}{|\mathcal{L}_{i} \cup \mathcal{L}_{i-1}|}$$

\begin{figure*}[ht!]
    \centering
    \begin{subfigure}{0.4\textwidth}
        \includegraphics[width=\textwidth]{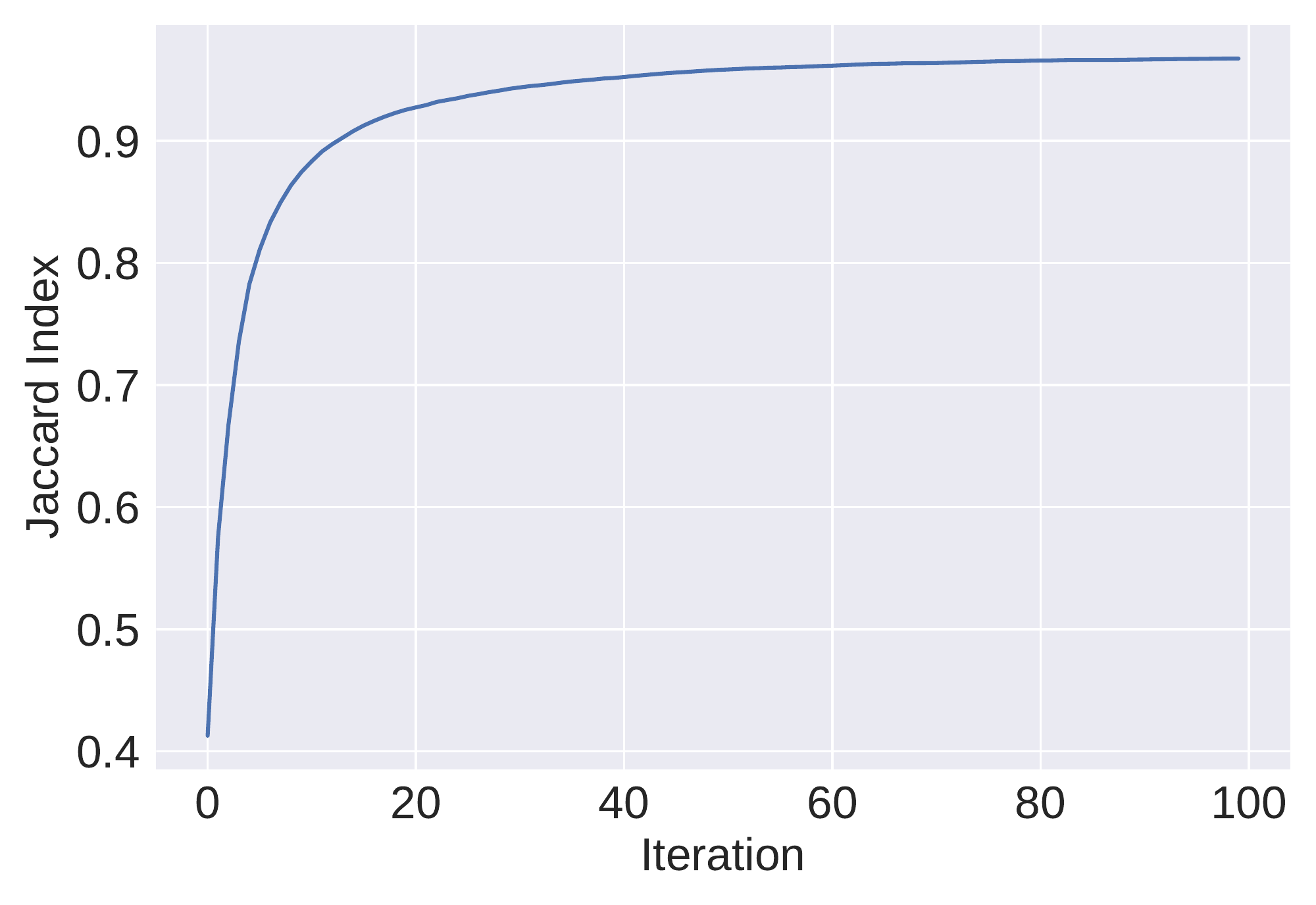}
        \caption{English (CCOHA).}
    \end{subfigure}
    \qquad
    \begin{subfigure}{0.4\textwidth}
        \includegraphics[width=\textwidth]{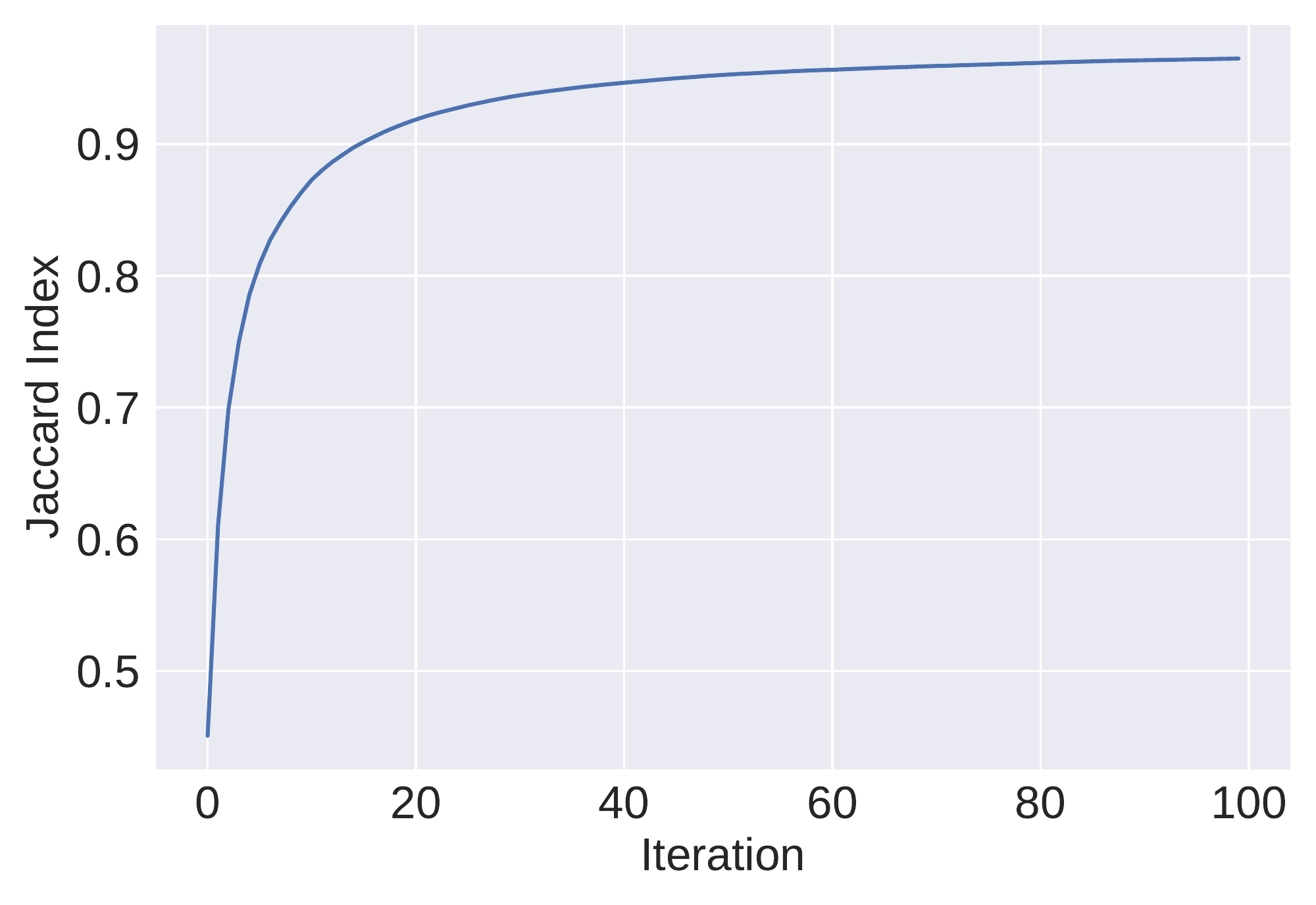}
        \caption{German (DTA+BZ+ND).}
    \end{subfigure}
    
        \begin{subfigure}{0.4\textwidth}
        \includegraphics[width=\textwidth]{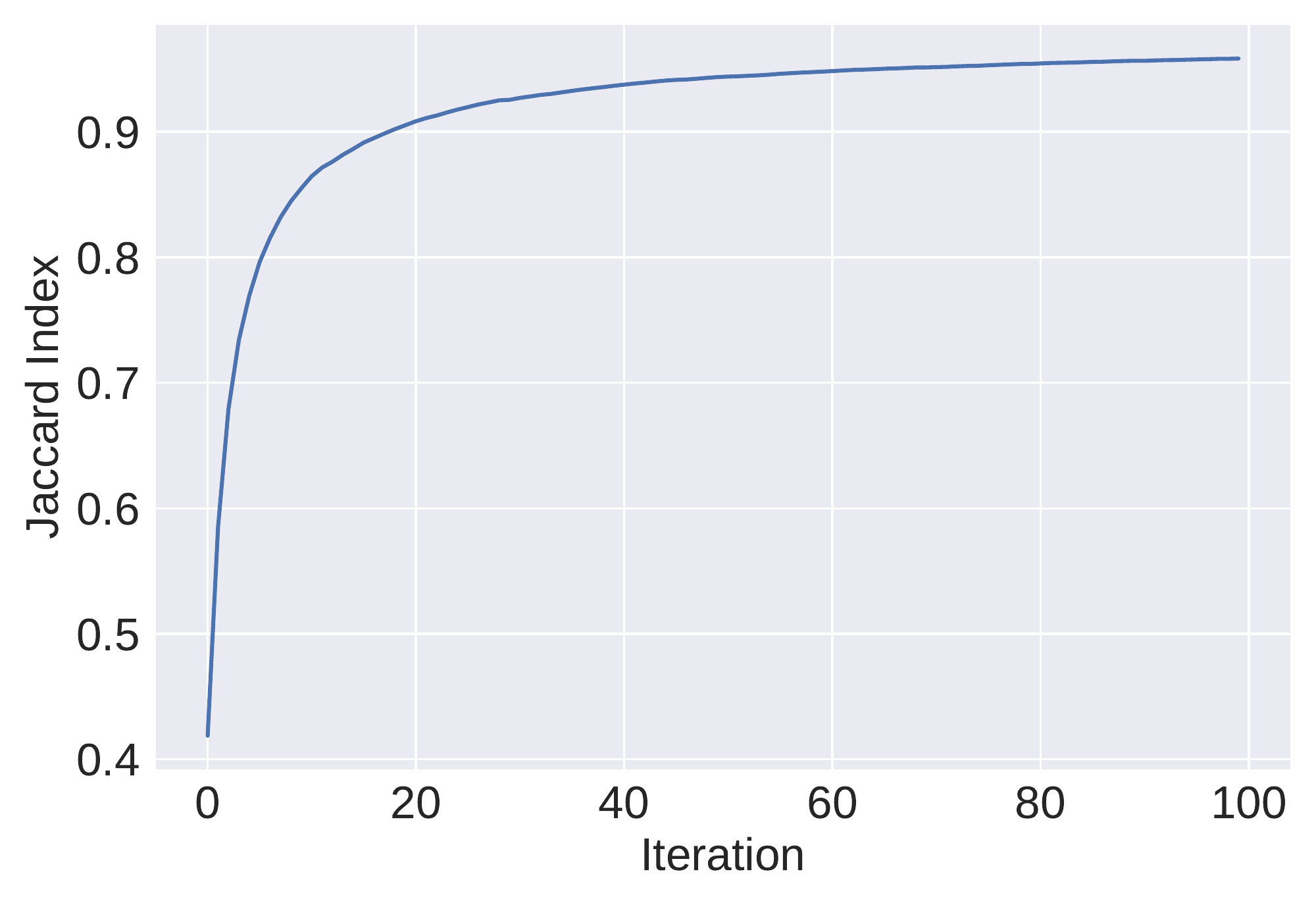}
        \caption{Latin (LatinISE).}
    \end{subfigure}
    \qquad
    \begin{subfigure}{0.4\textwidth}
        \includegraphics[width=\textwidth]{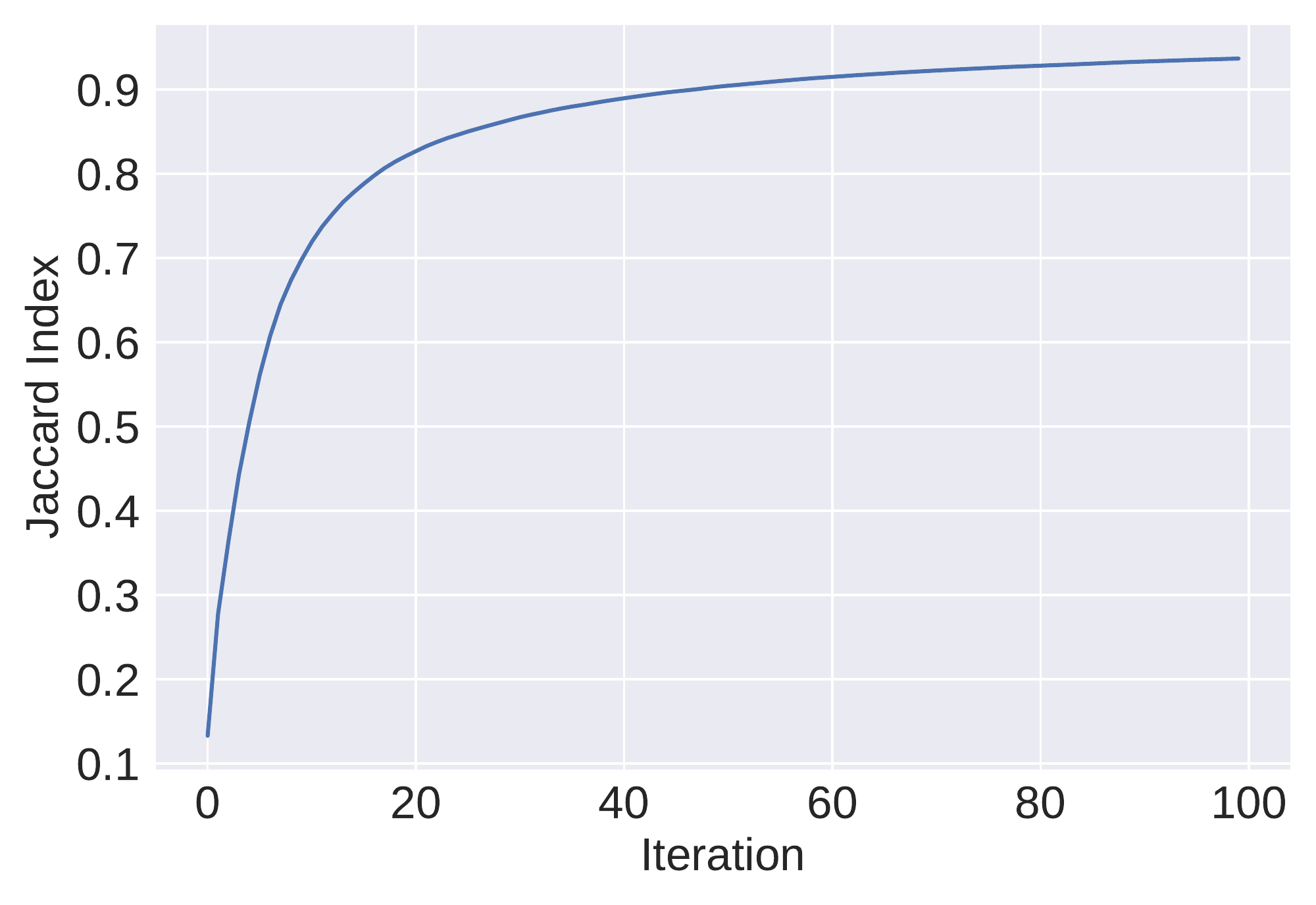}
        \caption{German (KubHist).}
    \end{subfigure}

    \caption{Overlap of landmark sets between iterations measured by the Jaccard Index. Landmark selection converges to a stable set after a few iterations.}
    \label{fig:overlap}
\end{figure*}

Figure \ref{fig:overlap} shows the running average of the Jaccard Index computed at each iteration for each language in this experiment. 
For every data set, the Jaccard Index is above 0.95 at 100 iterations, indicating a nearly perfect overlap between the chosen landmarks. 
This shows that \ourmethodA~ is not subject to catastrophic forgetting, being able to keep information from previous iterations.

\section{Semantic Shift Discovery (Section \ref{sec:discovery})}
\label{app:arxiv}

In this section we show additional results to the arXiv semantic shift discovery experiment. We test the word discovery between various arXiv subjects. Specifically, we show the unique discoveries made by Global, Noise-Aware and \ourmethodA~, as well as the common discoveries, among the top 50 most shifted words (i.e., most unstable words).

\paragraph{Artificial Intelligence vs. Classical Physics}
In addition to the unique words shown in Table \ref{tab:words} (Section \ref{sec:discovery}). These results are shown in Table \ref{tab:unique_ai_class}.

\begin{table*}[ht!]
    \centering
    \caption{Unique Discoveries: semantic shifts uniquely discovered by each alignment method among the 50 most shifted for Artificial Intelligence vs. Classical Physics. Common Words: top 20 most shifted words commonly predicted by all three alignment methods.}
    \begin{tabular}{ccc|cc}
        \multicolumn{3}{c}{\textbf{Unique Discoveries}} & \multicolumn{2}{c}{\textbf{Common Words}} \\
        \textbf{Global} & \textbf{Noise-Aware} & \multicolumn{1}{c}{\textbf{\ourmethodA}} &  \\ \toprule
        agent &  & components & concepts & nodes \\
        approximation &  & element & density & phys \\
        boundary &  & mass & deterministic & polynomial \\
        conceptual &  & order & die & probability \\
        knowledge &  & solution & edge & respect \\
        plane &  & space & equations & rev \\
        reference &  & state & fields & rough \\
        rules &  & term & internal & rule \\
        system &  & time & light & tensor \\
        systems &  & vector & los & variables \\
    \end{tabular}
    \label{tab:unique_ai_class}
\end{table*}

\paragraph{Artificial Intelligence vs. Material Sciences}
We compare arXiv subjects cs.AI and cond-mat.mtrl-sci. The size of the common vocabulary is 2214 words. Table \ref{tab:unique_ai_mtrl} shows the unique most shifted words for each method.

\begin{table*}[ht!]
    \centering
    \caption{Unique Discoveries: semantic shifts uniquely discovered by each alignment method among the 50 most shifted for Artificial Intelligence vs. Material Sciences. Common Words: words commonly discovered by all three alignment methods.}
    \begin{tabular}{ccc|cc}
        \multicolumn{3}{c}{\textbf{Unique Discoveries}} & \multicolumn{2}{c}{\textbf{Common Words}} \\
        \textbf{Global} & \textbf{Noise-Aware} & \multicolumn{1}{c}{\textbf{\ourmethodA}} &  \\ \toprule
        artificial & fields & atoms & approximation & phase \\
        conceptual & rule & energy & concepts & phys \\
        direction &  & functions & edge & probability \\
        exchange &  & rate & eqs & quantum \\
        field &  & relation & equations & quot \\
        knowledge &  & state & estimates & rough \\
        measurements &  & states & functional & space \\
        solving &  & theorem & logic & spin \\
         &  &  & lower & symmetry \\
         &  &  & monte & variables \\
    \end{tabular}
    \label{tab:unique_ai_mtrl}
\end{table*}

\paragraph{General Mathematics vs. Atomic Physics}
We compare arXiv subjects math.GM and physics.atom-ph. The size of the common vocabulary is 1950 words. Table \ref{tab:unique_gm_atom} shows the unique most shifted words for each method.

\begin{table*}[ht!]
    \centering
    \caption{Unique Discoveries: semantic shifts uniquely discovered by each alignment method among the 50 most shifted for General Mathematics vs. Atomic Physics. Common Words: words commonly discovered by all three alignment methods.}
    \begin{tabular}{ccc|cc}
        \multicolumn{3}{c}{\textbf{Unique Discoveries}} & \multicolumn{2}{c}{\textbf{Common Words}} \\
        \textbf{Global} & \textbf{Noise-Aware} & \multicolumn{1}{c}{\textbf{\ourmethodA}} &  \\ \toprule
        appl & positive & closed & basis & mass \\
        continuous &  & element & configuration & math \\
        correlation &  & function & contributions & measure \\
        data &  & functions & differential & moment \\
        expectation &  & matrix & ground & quantum \\
        fine &  & non & ideal & sequence \\
        grant &  & numbers & infinity & series \\
        group &  & open & interpretation & transitions \\
        perturbation &  & right & know & uncertainty \\
        points &  & set & loop & wave \\
        problem &  & sets &  &  \\
        sequences &  & space &  &  \\
        spaces &  & table &  &  \\
        variation &  & value &  &  \\
    \end{tabular}
    \label{tab:unique_gm_atom}
\end{table*}

\subsection{Order of Input Corpora}

In this section, we investigate the qualitative differences between the discovered semantic changes in inputs $A$ and $B$ when inverting the order of input, that is, first when the input is $A, B$ and we simulate semantic change in $B$, then when the input is $B, A$ and we simulate the semantic change in $A$ for the self-supervision. No difference in the output was noticed under these circumstances. This implies that our method is stable with respect to the order of the input, revealing the same set of semantically changed words regardless of the input order.

\section{Additional Word Embeddings}
\label{app:emb}

In this section we report results for S4 using GloVe \cite{pennington2014glove} and FastText \cite{bojanowski2017enriching}. GloVe, like Word2Vec, creates word embeddings based on the distributional semantics property alone, whereas FastText includes sub-word information in the training objective, clustering words according to their lexical similarity. Our goal is to investigate whether the sub-word information can be used in conjunction with our self-supervised method for detecting semantic change. The parameters used for training the embeddings were window size $10$, sub-word length $4$ (fast text only), the vector dimension was $300$.

We repeat the experiments from Section \ref{sec:ukus} and Section \ref{sec:semeval}, results are reported in Tables \ref{tab:glove_ukus}, \ref{tab:ft_ukus}, \ref{tab:glove_semeval}, \ref{tab:ft_semeval}. The GloVe embeddings do not show a decrease in accuracy as high as the FastText ones.
For FastText, the results show a decrease in accuracy in nearly every situation, this is likely due to a conflict in the assumptions of semantic change and the use of sub-word information. Contextual word embeddings such as Word2Vec directly encodes the distributional semantics property. However, sub-word information adds signals unrelated to the contextual distribution, thus creating noisy clusters of words with respect to neighborhood. Since our semantic change detection is based in the distributional semantics property, we may expect loss in performance when adding sub-word information.

\begin{table*}[ht]
    \centering
    \caption{Results for the British vs. American English experiment using GloVe.
    }
    \begin{tabular}{llrrrr}
          \textbf{Method}                & \textbf{Alignment} & \textbf{Accuracy} & \textbf{Precision} & \textbf{Recall} & \textbf{F1} \\ \toprule
          COS                     & \multirow{2}{*}{Global}  & $0.36/0.28/0.24$ & $0.46/0.34/0.00$ & $0.27/0.10/0.00$ & $0.33/0.14/0.00$ \\
          \ourmethodD                 &  & $0.44$ $\pm 0.04$ & 0.78 $\pm 0.05$ & $0.29 \pm 0.06$ & 0.44 $\pm 0.06$ \\   \midrule    
          COS                        & \multirow{2}{*}{Top-5\%} & $0.38/0.29/0.24$ & $0.44/0.33/0.00$ & $0.27/0.11/0.00$ & $0.31/0.14/0.00$ \\
          \ourmethodD                 &  & 0.65 $\pm 0.03$ & $0.80 \pm 0.02$ & $0.78 \pm 0.03$ & 0.82 $\pm 0.01$ \\ \midrule
          COS                        & \multirow{2}{*}{Top-10\%} & $0.35/0.30/0.24$ & $0.45/0.33/0.00$ & $0.22/0.08/0.00$ & $0.29/0.13/0.00$ \\
          \ourmethodD                 &  & $0.68 \pm 0.03$ & $0.81 \pm 0.03$ & $0.74 \pm 0.04$ & $0.74 \pm 0.03$\\ \midrule
          COS                        & \multirow{2}{*}{Bot-5\%} & $0.38/0.28/0.25$ & $0.45/0.30/0.00$ & $0.30/0.09/0.00$ & $0.34/0.14/0.00$ \\
          \ourmethodD                 & & $0.43 \pm 0.03$ & $0.61 \pm 0.02$ & $0.20 \pm 0.03$ & $0.30 \pm 0.02$ \\ \midrule
          COS                        & \multirow{2}{*}{Bot-10\%} & $0.38/0.28/0.24$ & $0.51/0.42/0.22$ & $0.26/0.09/0.03$ & $0.33/0.16/0.02$ \\
          \ourmethodD                 &  & $0.62 \pm 0.03$ & $0.70 \pm 0.02$ & $0.87 \pm 0.03$ & $0.77 \pm 0.02$ \\ \midrule
          COS        & \ourmethodA       & $0.42/0.34/0.29$  & $0.63/0.59/0.52$ & $0.38/0.15/0.05$  & $0.41/0.22/0.11$ \\ 
          \ourmethodD    & \ourmethodA   & $0.66 \pm 0.01$  & $0.70 \pm 0.02$  & $0.90 \pm 0.02$   & $0.78 \pm 0.03 $\\ \midrule
          Noisy-Pairs                & -  & $0.27$ & $\mathbf{1.00}$ & $0.04$ & $0.07$ \\ \bottomrule

     \end{tabular}
    \label{tab:glove_ukus}
\end{table*}

\begin{table*}[ht]
    \centering
    \caption{Results for the British vs. American English experiment using FastText.
    }
    \begin{tabular}{llrrrr}
         \textbf{Method}                & \textbf{Alignment} & \textbf{Accuracy} & \textbf{Precision} & \textbf{Recall} & \textbf{F1} \\ \toprule
         COS                     & \multirow{2}{*}{Global}  & $0.29/0.20/0.19$ & $0.40/0.20/0.00$ & $0.20/0.08/0.00$ & $0.22/0.13/0.00$ \\
         \ourmethodD                 &  & $0.40 \pm 0.05$ & 0.78 $\pm 0.05$ & $0.22 \pm 0.04$ & 0.41 $\pm 0.06$ \\   \midrule    
         COS                        & \multirow{2}{*}{Top-5\%} & $0.30/0.22/0.16$ & $0.32/0.20/0.00$ & $0.22/0.09/0.00$ & $0.22/0.10/0.00$ \\
         \ourmethodD                 &  & 0.49 $\pm 0.02$ & $0.70 \pm 0.02$ & $0.65 \pm 0.03$ & 0.67 $\pm 0.02$ \\ \midrule
         COS                        & \multirow{2}{*}{Top-10\%} & $0.32/0.27/0.21$ & $0.41/0.29/0.00$ & $0.22/0.08/0.00$ & $0.29/0.10/0.00$ \\
         \ourmethodD                 &  & $0.52 \pm 0.04$ & $0.66 \pm 0.03$ & $0.68 \pm 0.02$ & $0.69 \pm 0.01$\\ \midrule
         COS                        & \multirow{2}{*}{Bot-5\%} & $0.22/0.19/0.10$ & $0.41/0.28/0.00$ & $0.29/0.08/0.00$ & $0.30/0.08/0.00$ \\
         \ourmethodD                 & & $0.40 \pm 0.03$ & $0.59 \pm 0.03$ & $0.17 \pm 0.03$ & $0.27 \pm 0.02$ \\ \midrule
         COS                        & \multirow{2}{*}{Bot-10\%} & $0.32/0.27/0.22$ & $0.40/0.22/0.16$ & $0.27/0.10/0.01$ & $0.31/0.15/0.01$ \\
         \ourmethodD                 &  & $0.59 \pm 0.04$ & $0.68 \pm 0.02$ & $0.81 \pm 0.03$ & $0.77 \pm 0.02$ \\ \midrule
         COS        & \ourmethodA       & $0.32/0.26/0.21$  & $0.41/0.32/0.28$ & $0.14/0.08/0.02$  & $0.11/0.05/0.02$ \\ 
         \ourmethodD    & \ourmethodA   & $0.51 \pm 0.02$  & $0.60 \pm 0.01$  & $0.50 \pm 0.02$   & $0.58 \pm 0.01 $\\ \midrule
         Noisy-Pairs                & -  & $0.22$ & $\mathbf{1.00}$ & $0.04$ & $0.06$ \\ \bottomrule

    \end{tabular}
    \label{tab:ft_ukus}
\end{table*}

\begin{table}[ht]
    \centering
    \caption{Results for the SemEval-2020 experiment using GloVe word embeddings.}
    \begin{tabular}{lrrrr}
         \textbf{Alignment} & \multicolumn{1}{c}{\textbf{English}} & \multicolumn{1}{c}{\textbf{German}} & \multicolumn{1}{c}{\textbf{Latin}} & \multicolumn{1}{c}{\textbf{Swedish}}  \\ \toprule

         \ourmethodA     & 0.68 & 0.79 & 0.65 & 0.77 \\
         Noise-Aware    & 0.65 & 0.75 & 0.65 & 0.74 \\
         Top 5\% fr.    & 0.65 & 0.77 & 0.68 & 0.77 \\
         Top 10\% fr.   & 0.65 & 0.79 & 0.68 & 0.74 \\
         Bot 5\% fr.    & 0.62 & 0.73 & 0.62 & 0.77 \\
         Bot 10\% fr.   & 0.62 & 0.75 & 0.68 & 0.74 \\
         Global         & 0.65 & 0.73 & 0.57 & 0.74 \\ \midrule
        SemEval \#1 & \multicolumn{1}{r}{\textbf{0.70}} & \multicolumn{1}{r}{0.79} & \multicolumn{1}{r}{0.68} & \multicolumn{1}{r}{\textbf{0.81}} \\
        SemEval \#2 & \multicolumn{1}{r}{\textbf{0.70}} & \multicolumn{1}{r}{0.77} & \multicolumn{1}{r}{0.72} & \multicolumn{1}{r}{0.74} \\
        SemEval \#3 & \multicolumn{1}{r}{0.65} & \multicolumn{1}{r}{0.77} & \multicolumn{1}{r}{0.75} & \multicolumn{1}{r}{0.74} \\
        \bottomrule
         
    \end{tabular}
    \label{tab:glove_semeval}
\end{table}

\begin{table}[ht]
    \centering
    \caption{Results for the SemEval-2020 experiment using FastText word embeddings.}
    \begin{tabular}{lrrrr}
         \textbf{Alignment} & \multicolumn{1}{c}{\textbf{English}} & \multicolumn{1}{c}{\textbf{German}} & \multicolumn{1}{c}{\textbf{Latin}} & \multicolumn{1}{c}{\textbf{Swedish}}  \\ \toprule

         \ourmethodA     & 0.65 & 0.77 & 0.62 & 0.74 \\
         Noise-Aware    & 0.65 & 0.75 & 0.65 & 0.74 \\
         Top 5\% fr.    & 0.65 & 0.77 & 0.68 & 0.77 \\
         Top 10\% fr.   & 0.65 & 0.79 & 0.68 & 0.74 \\
         Bot 5\% fr.    & 0.57 & 0.73 & 0.62 & 0.77 \\
         Bot 10\% fr.   & 0.62 & 0.75 & 0.68 & 0.74 \\
         Global         & 0.65 & 0.73 & 0.57 & 0.74 \\ \midrule
        SemEval \#1 & \multicolumn{1}{r}{\textbf{0.70}} & \multicolumn{1}{r}{0.79} & \multicolumn{1}{r}{0.68} & \multicolumn{1}{r}{\textbf{0.81}} \\
        SemEval \#2 & \multicolumn{1}{r}{\textbf{0.70}} & \multicolumn{1}{r}{0.77} & \multicolumn{1}{r}{0.72} & \multicolumn{1}{r}{0.74} \\
        SemEval \#3 & \multicolumn{1}{r}{0.65} & \multicolumn{1}{r}{0.77} & \multicolumn{1}{r}{0.75} & \multicolumn{1}{r}{0.74} \\
        \bottomrule
         
    \end{tabular}
    \label{tab:ft_semeval}
\end{table}

\section{Code}

The implementation of the methods and experiments in this paper are provided as part of the supplementary material. The code is written in Python3, and uses Tensorflow2, and Keras to build the machine learning models. Further details about dependencies is available in the code package.

\end{document}